\documentclass[10pt,journal,compsoc]{IEEEtran}

\ifCLASSOPTIONcompsoc
  \usepackage[nocompress]{cite}
\else
  \usepackage{cite}
\fi
\ifCLASSINFOpdf
\else
\fi
\usepackage{times}  
\usepackage{helvet}  
\usepackage{courier}  
\usepackage{url}  
\usepackage{graphicx} 
\usepackage{caption} 

\usepackage{algorithm}
\usepackage{algorithmic}
\usepackage[]{xcolor}

\usepackage{newfloat}
\usepackage{listings}

\usepackage{amsmath}
\usepackage{multirow}
\usepackage{amssymb}
\usepackage{bbm}
\usepackage{dsfont}
\usepackage{graphicx}
\usepackage{cleveref}
\usepackage{subcaption}
\usepackage{caption}
\usepackage{colortbl}
\usepackage{array}
\newcolumntype{P}[1]{>{\centering\arraybackslash}p{#1}}
\begin{document}
%
\title{\LARGE Learning to Adapt to Online Streams with Distribution Shifts}

\author{Chenyan Wu, Yimu Pan, Yandong Li, James Z. Wang,~\IEEEmembership{Senior Member,~IEEE}
\IEEEcompsocitemizethanks{\IEEEcompsocthanksitem C. Wu, Y. Pan, and J. Z. Wang are with the Data Science and Artificial Intelligence Area of the College of Information Sciences and Technology, The Pennsylvania State University, University Park, PA 16802, USA
(e-mails: \{czw390, ymp5078, jwang\}@psu.edu).
\IEEEcompsocthanksitem Y. Li is with Google Research, Bellevue, WA 98004, USA
(e-mail: lyndon.leeseu@outlook.com).
\IEEEcompsocthanksitem (Corresponding author: C. Wu.)
}
}
\IEEEtitleabstractindextext{%
\begin{abstract}
Test-time adaptation (TTA) is a technique used to reduce distribution gaps between the training and testing sets by leveraging unlabeled test data during inference. In this work, 
we expand TTA to a more practical scenario, where the test data comes in the form of online streams that experience distribution shifts over time. 
Existing approaches face two challenges: reliance on a large test data batch from the same domain and the absence of explicitly modeling the continual distribution evolution process.
To address both challenges, we propose a meta-learning approach that teaches the network to adapt to distribution-shifting online streams during meta-training.
As a result, the trained model can perform continual adaptation to distribution shifts in testing, regardless of the batch size restriction, as it has learned during training.
We conducted extensive experiments on benchmarking datasets for TTA, incorporating a broad range of online distribution-shifting settings. Our results showed consistent improvements over state-of-the-art methods, indicating the effectiveness of our approach. In addition, we achieved superior performance in the video segmentation task, highlighting the potential of our method for real-world applications.
\end{abstract}

}

\maketitle

\IEEEdisplaynontitleabstractindextext

%
\IEEEpeerreviewmaketitle

\section{Introduction}\label{sec:introduction}
\IEEEPARstart{M}{achine} learning systems typically assume that training and testing datasets have similar distributions. However, the assumption often fails in practice.
The distribution shift between the training (source) and testing (target) datasets widely exists, which might degrade the model performance in testing~\cite{lazer2014parable, recht2018cifar}.
In recent years, test-time adaptation (TTA) methods, which aim at alleviating the distribution shift with only test data, have gained popularity~\cite{li2016revisiting,sun2020test,lipton2018detecting, wang2020tent, zhang2021adaptive}. Unlike the regular domain adaptation setting, which allows the model to observe the data from the test distribution during training, TTA forbids such observation. The adjustment is more aligned with real-world scenarios as it is common to have limited knowledge about the target domains during the training phase.

\begin{figure}[!t]
  \begin{center}
    \includegraphics[trim=0 0 0 0, width=0.95\linewidth]{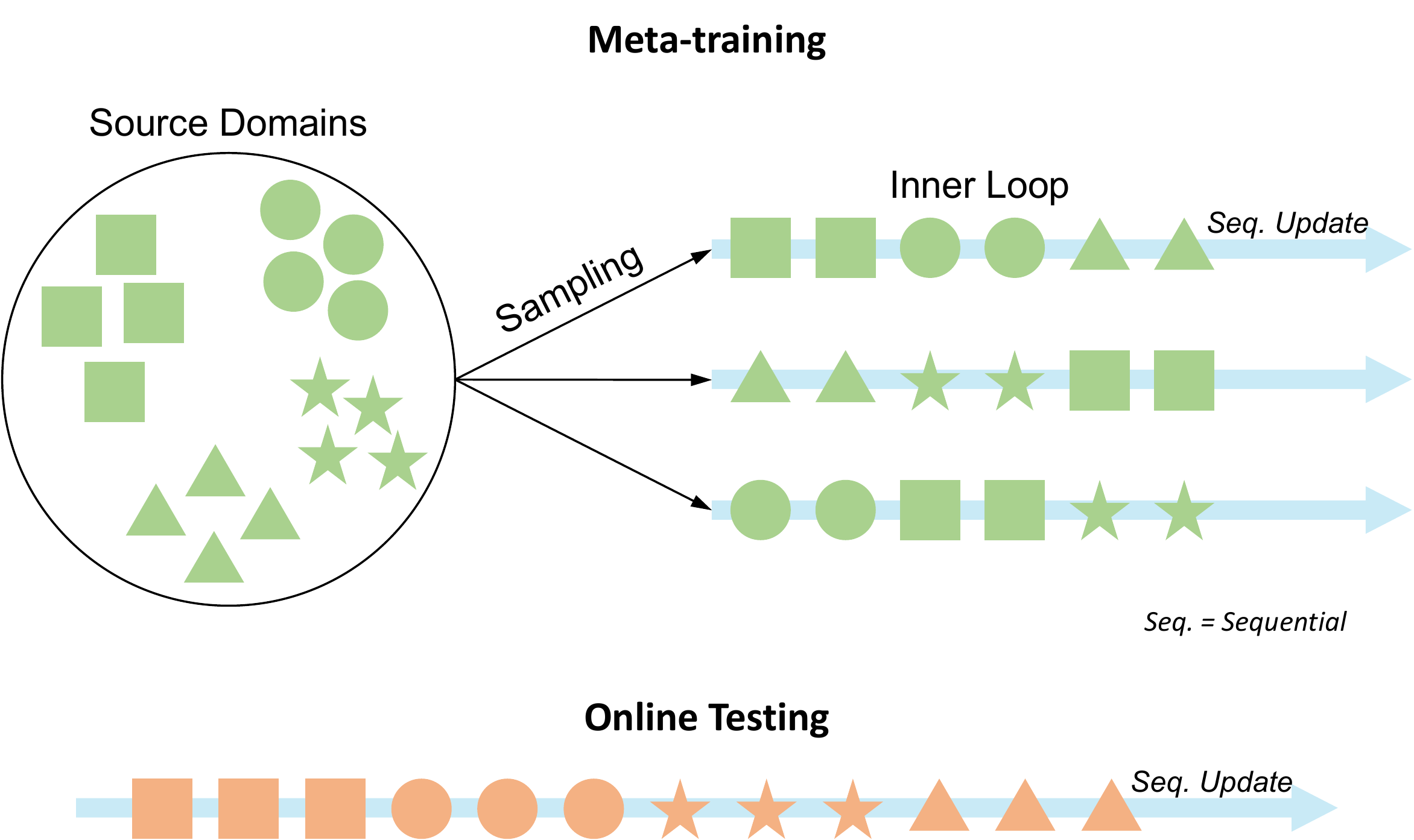}
  \end{center}
  \caption{Our method simulates the online testing procedure in the inner loop of meta-training. Objects with different shapes and colors represent samples from different domains. Of note, for better visualization, the settings depicted in the figure, including the number of source domains and the duration for each domain, differ from the experimental settings used in our work.}\label{fig:setting}
\end{figure}
Leveraging information from unlabeled test data, such as batch statistics~\cite{li2016revisiting,wang2020tent}, is crucial to TTA methods.
In this process, however, most existing approaches implicitly assume the test data all comes from the same stationary domain.
In practice, when data comes in an online manner, their distributions usually shift over time.
For example, the environment an autonomous vehicle operates in can be affected by the weather and lighting conditions, which constantly change over time.
Throughout this article, we refer to this type of data as distribution-shifting online streams. 
It is worth noting that the duration of each distribution might vary in the stream.
For instance, a dark rainy day can last for several hours, but lightning illuminates the sky for a few seconds.

The following two limitations could lead to failure when applying existing TTA methods to distribution-shifting online streams.
First, several methods~\cite{li2016revisiting,wang2020tent,zhang2021adaptive,wang2022continual} rely on optimizing a batch of test data to make the adaptation.
The batch should come from one domain and include sufficient data points to characterize its distribution.
However, as aforementioned, the duration of domains is unknown, which prevents us from guaranteeing that the data collected in a large batch belongs to a single domain.
These methods may not be robust to variations in domain duration and batch size.
We will validate this limitation in the experiment section. 
Second, existing work rarely explicitly models how to adapt to multiple domains continually. In our setting, the network needs to quickly adapt to the following domain after having already adapted to one domain. Some existing methods only involve stationary domains in training and testing~\cite{li2016revisiting,sun2020test,lipton2018detecting, wang2020tent}. Therefore, they have no chance to learn how to make such a continual adaptation.

We propose a novel meta-learning approach to address the limitations of adapting to distribution-shifting streams. Our approach explicitly trains the neural network to adapt to these streams by adopting the ``learning to learn" philosophy~\cite{hochreiter2001learning} and leveraging the MAML~\cite{finn2017model} framework as the training strategy. In the inner loop of meta-training, we sample data from multi-domain training data to generate streams and sequentially adapt the network to the stream samples using the self-supervised learning method, as shown in Fig.~\ref{fig:setting}. This process simulates online test-time adaptation for distribution-shifting streams. The outer loop enables the network to adapt to various domains after the sequential updates in the inner loop.
Hence, the meta-trained network can continually adapt to multiple domains on the online streams during testing.
Since our adaptation method optimizes stream samples, there is no batch size restriction.

We construct training data and distribution-shifting online testing streams using the datasets CIFAR-$10$-C~\cite{recht2018cifar} and Tiny ImageNet-C~\cite{hendrycks2019benchmarking}. 
We first demonstrate the limitations of some typical TTA methods~\cite{wang2020tent,sun2020test} when applied directly to those online streams, highlighting the need for a more effective approach.
Then we compare our method with state-of-the-art methods in a range of distribution-shifting settings, including periodic and randomized distribution shifts with varying domain durations.
Extensive experiments show that our method significantly outperforms state-of-the-art methods under these settings; in particular, our method introduces over $3\%$ test accuracy improvement than competing methods when the distribution-shifting duration is short.
We provide a comprehensive ablation analysis to demonstrate the effectiveness of each part of our method. Furthermore, we apply our method to the practical problem of video semantic segmentation, achieving significant performance improvement on the CamVid dataset~\cite{brostow2009semantic}, and highlighting the potential for real-world applications.

{Our contributions can be summarized as follows:
\begin{itemize}
\item In the context of real-world settings with distribution-shifting data, we identify two key limitations of current test-time online adaptation approaches: the reliance on batch optimization and a lack of explicitly modeling multiple continuous domains.
\item According to the two limitations, we propose a novel meta-learning framework that simulates online adaptation updates on distribution-shifting streams.
\item Through extensive experimentation, we validate the effectiveness of our method in a variety of distribution-shifting settings as well as a practical problem (i.e., video semantic segmentation). The results demonstrate the superiority of our method, particularly in streams with fast distribution shifts.
\end{itemize}}

\section{Related Work}
Test-time adaptation (TTA) in distribution-shifting streams is related to multiple research areas, although the setup is different. This section discusses related work from the areas of domain adaptation, meta-learning, and online learning.

\textit{Domain adaptation} is a well-studied topic that aims to address the distribution shift problem~\cite{gretton2009covariate, sun2017correlation, ganin2015unsupervised, tzeng2017adversarial, sun2019unsupervised}. The core idea behind these methods is to align features of the source and target data~\cite{gretton2009covariate, sun2017correlation}, for example, by adversarial invariance~\cite{ganin2015unsupervised, tzeng2017adversarial} or shared proxy tasks~\cite{sun2019unsupervised}. Most methods assume test data from target domains is available in the training stage. However, they only consider one fixed target domain in the training stage~\cite{shimodaira2000improving, daume2009frustratingly, gong2012geodesic,ganin2015unsupervised,tzeng2017adversarial}. As a result, it is difficult for them to handle streams where the target distribution changes dynamically.
Some recent studies apply a source-free manner to make domain adaptation~\cite{li2020model, kundu2020universal, liang2020we}, which focuses on training target data in the training stage. These methods all use complex training strategies. \cite{li2020model, kundu2020universal} adopt generative models of target data. \cite{liang2020we} uses entropy minimization and diversity regularization. These methods all need to optimize target data offline for several epochs. They cannot perform adaptation according to the coming test sample, namely the online test.
All the above methods only perform adaptation in training, so we call them train-time adaptation methods. 

There are several approaches to test-time unsupervised adaptation~\cite{li2016revisiting,sun2020test,lipton2018detecting, wang2020tent, zhang2021adaptive}, which are more related to our method. In particular, Li et al.~\cite{li2016revisiting} focus on the batch normalization layer for test-time adaptation. TENT~\cite{wang2020tent} applies entropy minimization to test time adaptation.  CoTTA\cite{wang2022continual} proposes an extension to TENT by producing pseudo-labels on each test sample using averaged results from many augmentations and restoring the student model parameters to prevent over-fitting to a specific domain.
ARM~\cite{zhang2021adaptive} focuses on group data where one group of data corresponds to one distribution. This work adopts the “learning to learn” idea, where the training strategy is similar to MAML~\cite{finn2017model}. However, it requires a batch of data, whereas our method works on one sample.
NOTE~\cite{gongnote} studies the streams where the samples are temporally correlated, namely, the classes of successive samples are related. However, it does not consider the shifts in distributions.

\textit{Meta-learning} is mostly used in few-shot learning~\cite{santoro2016meta, vinyals2016matching,ravi2016optimization, snell2017prototypical, finn2017model}. Some prior arts~\cite{li2018learning, dou2019domain} utilize meta-learning methods for domain generalization, which differs from our effort. \cite{zhang2021adaptive} apply meta-learning to TTA, but it focuses on groups of data instead of streams of data. We extend the meta-learning method to work on online stream adaptation. Specifically, our method extends MAML~\cite{finn2017model} algorithm in two different directions. 
Firstly, our method is unsupervised, using only unsupervised loss to optimize model weights in the inner loop instead of ground truth labels. 
Second, our method works on online streams to mimic the testing setting. In the inner loop, traditional methods update the model parameters based on only the loss of a batch of data, whereas our method optimizes over samples. 

\textit{Online learning} naturally handles stream data~\cite{shalev2011online, hazan2019introduction}. The SOTA framework generally repeats the following operations on each data: receive a sample, obtain a prediction, receive the label from a worst-case oracle, and update the model. 
Most online learning methods need ground-truth labels, which is unsuitable for our setting. 
Although the recent online learning approach~\cite{wu2021online} extends the test-time adaptation to the unsupervised setting.
It mainly focuses on the label distribution shift other than the distribution shift of data.

\section{Problem Setting}\label{section:problem}
We first introduce the generation of training and test data, followed by a description of the test-time adaptation model.

Let $\mathcal{P}$ be the data distribution defined over the space $\mathcal{X} \times \mathcal{Y}$, where $x \in \mathcal{X}$ and $y \in \mathcal{Y}$ represent the input and output, respectively. Our training dataset consists of samples from multiple distributions, denoted by the set $\mathcal{P}^S$ where $|S|$ denotes the number of distributions. The objective is for the model to learn from different distributions during training. The availability of datasets with multiple domains is readily feasible, as a variety of everyday conditions can result in differing data distributions. For instance, in the context of autonomous driving, images obtained at different time points or in varying weather conditions can be treated as distinct data domains.

During the testing time, we receive distribution-shifting online streams, where the data distribution changes over time.
We define the set of all test distributions as $\mathcal{P}^T$, which includes the distributions encountered during testing. 
In this test-time adaptation setting, $\mathcal{P}^S$ and $\mathcal{P}^T$ do not overlap.
Consider the test stream $\mathcal{T}$ in a discrete-time scale, 
\begin{equation}
  \mathcal{T} = (x_1, y_1),(x_2, y_2),...,(x_t, y_t),...\;.
\end{equation}
At the time step $t$, $(x_t, y_t)\sim \mathcal{P}_t$, where $\mathcal{P}_t \in \mathcal{P}^T$ and $\mathcal{P}_t$ changes over time.
Since data over a period of time often share the same distribution, we divide the time scale into time periods $1,...,T_1-1,T_1,..., T_2-1,T_2,...$, where $(x_t, y_t)\sim \mathcal{P}_{T_i}$ for any $t \in (T_{i-1},T_i]$.
The duration of each period $(T_{i-1},T_i]$, can vary in length;
this setting mimics real-world scenarios, where a single domain might persist for a very long or very short period of time.
For example, a night lasts for hours, but a lightning strike only temporarily illuminates the sky.

Our goal is to train a model on the generated training data that adapts well to test streams. 
In the training stage, we aim to train a network $f(\cdot;\theta_0)$ on parameter $\theta_0$ using all the training data without constraint.
In the testing stage, we want the network to optimize $\theta_t$ using only the previous step parameter $\theta_{t-1}$ and the unlabeled current step input $x_t$, at each time step $t$. 
In other words, \textit{the network has no access to past data points, future data points, or training data in the testing stage}. 
We set such a constraint in the testing time since the distribution-shifting online stream arrives one sample at a time.

To evaluate the accuracy on the stream $\mathcal{T}$, we average the accuracy over all samples in $\mathcal{T}$, denoted as $\frac{1}{L} \sum^{L}_{t=1}\mathds{1} (f(x_t;\theta_t)=y_t)$, where $L$ is the length of $\mathcal{T}$. 

\begin{figure*}[ht]
  \begin{center}
    \includegraphics[trim=0 0 0 0, width=0.9\textwidth]{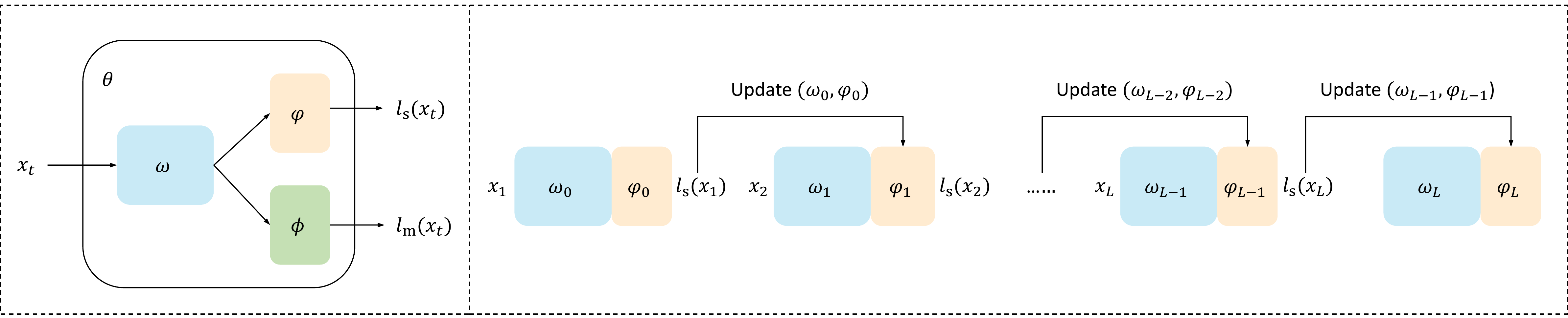}
  \end{center}
  \caption{The left figure illustrates the model structure. The model parameter $\theta$ is split into three parts: $\omega$ of the feature extractor, $\varphi$ of the self-supervised branch, and $\phi$ of the supervised branch. The right figure illustrates the online update procedure for the inner loop and the testing procedure.}\label{fig:framework}
\end{figure*}

\section{Method}
This section introduces our technical approach.
We explicitly guide the model to adapt to the distribution-shifting online streams using the meta-learning approach to tackle the defined problem in the previous section.
Our approach is inspired by the meta-learning philosophy of ``learning to learn'' and some recent studies that demonstrate the meta-learning's ability to quickly adapt to the stream data~\cite{nagabandi2018deep, javed2019meta, al2017continuous}. 
In particular, we adopt the typical meta-learning framework, MAML~\cite{finn2017model}, as our training strategy. 
We present the training framework in Algorithm~\ref{algorithm:framework}.
Similar to MAML, our framework consists of the inner loop (lines 3-7) and the outer loop (lines 8-10). 
In the inner loop, we first sample a distribution-shifting online stream $\mathcal{T}$ from the training distribution set $\mathcal{P}^S$, as the training data (line 3). 
Next, we sequentially update the parameters $\theta_t$ from $\theta_{t-1}$ on each sample $x_t$, using the self-supervised loss $l_\text{s}(f(x_t;\theta_{t-1}))$ (lines 4-7). 
In contrast to the typical MAML-based approaches that train on batched data in the inner loop using ground truth labels, our method takes advantage of the distribution-shifting online streams to mimic the testing procedure for better performance.
In the outer loop, we sample a batch of data from $\mathcal{P}^S$ (line 8) and optimize the $\theta_0$ using the supervised loss on the batch (lines 9-10). This batch of data is referred to as the support set in meta-learning. 
The testing procedure is the same as the inner loop of the training procedure. We apply the self-supervised loss $l_\text{s}$ to sequentially adapt the model to each sample of the test stream.
We detail training and testing procedures in the following subsections.
\begin{algorithm}[t!]
  \caption{The framework for meta-training}\label{algorithm:framework}
  \textbf{Require}: $\mathcal{P}^S$: Source domains
  \begin{algorithmic}[1] 
  \STATE Initialize $\theta_0$
  \WHILE{not done}
    \STATE Sample stream $\mathcal{T}=(x_1,x_2,...,x_L)$ from $\mathcal{P}^S$
    \FOR {$t=1,2...L$}
      \STATE Compute self-supervised loss \\
      $\mathcal{L}_\text{s}= l_\text{s}(f(x_t;\theta_{t-1}))$
      \STATE $\theta_{t} = \theta_{t-1} - \alpha\Delta_{\theta_{t-1}}\mathcal{L}_\text{s}$
    \ENDFOR
    \STATE Collect support set $\{({x}_i, {y}_i)\}_{i=1}^N$ from $\mathcal{P}^S$
    \STATE Compute supervised loss  \\
    $\mathcal{L}_\text{out} = \sum_{i=1}^{N}l_\text{out}(f(x_i;\theta_{L}), y_i)$
    \STATE $\theta_0  \leftarrow \theta_0 - \gamma\Delta_{\theta_0}\mathcal{L}_\text{out}$
  \ENDWHILE
  \end{algorithmic}
  \end{algorithm}

\subsection{Training Procedure}
\textit{Inner loop.} The inner loop mimics the test-time adaptation process for the distribution-shifting online streams. 
First, we construct the distribution-shifting online streams from the train set. 
We randomly sample $D$ distributions $\mathcal{P}_1,..., \mathcal{P}_i,...,\mathcal{P}_D$ in a sequence from the set $\mathcal{P}^S$.
Then $K$ data points are randomly sampled from each distribution to form a stream, $\mathcal{T}= (x_1,... x_2,..., x_t,...x_L)$, with the length $L=KD$, where $(x_t, y_t) \sim \mathcal{P}_{i}$ for all $t \in ((i-1)K, iK]$.
Because the model in the test-time adaptation setting has no access to the ground truth labels, we simulate this setting in the inner loop during training.
This simulation is done by applying the same self-supervised method both in the inner loop and in the testing stage. 
More specifically, we adopt the Test-Time Training (TTT) approach~\cite{sun2020test} to perform the self-supervised task of predicting image rotation angle. This task involves rotating an image by one of $0$, $90$, $180$, or $270$ degrees and classifying the angle of rotation as a four-way classification problem. The TTT approach, which operates on individual samples, bypasses the batch size requirement and provides a suitable solution for this problem.
To incorporate TTT into our method, we adopt a dual-branch structure, where the model parameter $\theta$ is split into three parts, $\omega$, $\varphi$ , and $\phi$ (see Fig.~\ref{fig:framework}).
The subnet $f(\cdot;\omega, \phi)$ is used to compute the supervised loss $l_\text{m}$ in the outer loop, whereas the self-supervised subnet $f(\cdot;\omega, \varphi)$ is used in the inner loop to compute $l_\text{s}$ in both the inner and the outer loop.
To start this inner loop, we initial parameters $\theta_0 = (\omega_0, \varphi_0 , \phi_0)$.
At time step $t$, with the current parameters $(\omega_{t-1}, \varphi_{t-1})$, we compute the self-supervised loss 
\begin{equation}\label{equ:inner1}
  \mathcal{L}_\text{s}=l_\text{s}(f(x_t;\omega_{t-1}, \varphi_{t-1})) \;.
\end{equation}
Then, $(\omega_{t-1}, \varphi_{t-1})$ are updated to $(\omega_t, \varphi_t)$ with $\mathcal{L}_\text{s}$. Formally, we have
\begin{equation}\label{equ:inner2}
  (\omega_{t}, \varphi_{t}) = (\omega_{t-1}, \varphi_{t-1}) - \alpha\Delta_{\omega_{t-1}, \varphi_{t-1}}\mathcal{L}_\text{s} \;,
\end{equation}
where $\alpha$ is the inner loop learning rate. \Cref{equ:inner1,equ:inner2} correspond to lines 5 and 6 in \Cref{algorithm:framework}, respectively. 
We continuously update the $(\omega, \varphi)$ along the stream $\mathcal{T}$ without modifying $\phi$ in the inner loop. Feeding a stream $\mathcal{T}$ with length $L$ into the model, we update $\theta_0$ by $L$ times and obtain $\theta_{L} = (\omega_L, \varphi_L , \phi_0)$ after one path over the inner loop. The whole inner loop update procedure is illustrated in Fig.~\ref{fig:framework}.

\textit{Outer loop.} 
After the self-supervised inner loop update, we hope the model can rapidly adapt to both the distributions that appeared in the inner loop and novel distributions.
Therefore, when constructing the support set, we keep the $D$ distributions, $\mathcal{P}_1,...,\mathcal{P}_D$, from the inner loop and randomly select additional $D^\prime$ distributions, $\mathcal{P}_{D+1},...,\mathcal{P}_{D+D^\prime}$, from $\mathcal{P}_S$.
Further, we randomly select $K$ samples from each of the $D$ domains, and one sample from each additional domain. Please note that a new set of $KD$ samples is selected from the $D$ domains, distinct from the samples in the inner loop. The resulting support set contains a total of $N=KD+D^\prime$ samples, denoted as ${{(x_i,y_i)}_{i=1}^N}$. The ablation analysis of the support set construction is presented in the experimental section.
In contrast to the inner loop, we also utilize ground truth labels $y_i$ to compute the supervised loss (i.e., cross-entropy) $l_\text{m}$ on $x_i$. 
We jointly train the supervised subnet $f(\cdot;\omega, \phi)$ and the self-supervised subnet $f(\cdot;\omega, \varphi)$ with $l_\text{m}$ and $l_\text{s}$, respectively. 
Using the parameters $\theta_{L} = (\omega_L, \varphi_L , \phi_0)$ obtained from the inner loop, we compute the total loss as
\begin{equation}\label{equ:out1}
  \mathcal{L}_\text{out}=\frac{1}{N}\sum^{N}_{i=1}l_\text{m}(f(x_i;\omega_L, \phi_0), y_i)+ l_\text{s}(f(x_i;\omega_L, \varphi_L))\;.
\end{equation}
Then, we compute the gradients to update $(\omega_0, \varphi_0 , \phi_0)$ using
\begin{equation}\label{equ:out2}
  (\omega_0, \varphi_0 , \phi_0)  \leftarrow (\omega_0, \varphi_0 , \phi_0) - \gamma\Delta_{\omega_0, \varphi_0 , \phi_0}\mathcal{L}_\text{out}\;,
\end{equation}
where $\gamma$ is the outer loop learning rate.
\Cref{equ:out1} corresponds to line 9 in \Cref{algorithm:framework},
where $l_\text{out}$ is the combination of $l_\text{m}$ and $l_\text{s}$.
\Cref{equ:out2} corresponds to line 10 in \Cref{algorithm:framework}.
\Cref{equ:out2} is a simplified representation of an optimization algorithm, and a more advanced optimizer like Adam can be adopted. 
It is worth noting that although we use $\theta_{L} = (\omega_L, \varphi_L , \phi_0)$ to compute $\mathcal{L}_\text{out}$, only the parameters $\theta_{0} = (\omega_0, \varphi_0 , \phi_0)$ are updated in the outer loop.
\subsection{Testing Procedure}\label{subsection:test}
The test data is in the form of the distribution-shifting online stream $(x_1, x_2, ..., x_t, ...)$. 
The online update procedure in testing is the same as the inner loop in training.
We start the online adaptation with the initial parameters $\theta_0 = (\omega_0, \varphi_0 , \phi_0)$ obtained from the training stage.
At time step $t$, $(\omega_{t-1}, \varphi_{t-1})$ are updated into $(\omega_{t}, \varphi_{t})$ by minimizing $l_\text{s}(f(x_t;\omega_{t-1}, \varphi_{t-1})$.
This optimization is the same as \Cref{equ:inner2}, whereas we use the test-time learning rate $\beta$ to replace $\alpha$.
Finally, we predict on $x_t$ using $f(x_t;\omega_t, \phi_0)$.

\section{Experiments}
The experiment section is organized as follows.
We first describe the datasets we use in this test-time adaptation setting, then list the baseline methods we compare to in this study.
Next, we perform confirmatory experiments to demonstrate the limitations of some existing methods.
Then, we conduct extensive experiments in different distribution-shifting settings and on different datasets to compare our method to others.
Additionally, we present ablation studies to provide a deeper understanding of our approach.
Finally, we conduct our method on a practical problem (i.e., video semantic segmentation).

\subsection{Datasets}\label{subsection:dataset}
We conduct the test-time adaptation experiments using the CIFAR-$10$-C and Tiny-Imagenet-C datasets~\cite{hendrycks2019benchmarking}.
Hendrycks et al.~\cite{hendrycks2019benchmarking} apply $15$ corruptions on the test sets of CIFAR-$10$ and Tiny ImageNet to construct the new datasets with different distributions, namely, CIFAR-$10$-C and Tiny ImageNet-C.
These corruptions (including noise, blur, weather, etc.) simulate the different environments in the real world. 
Each corruption has five severity levels, increasing from level $1$ to level $5$.
One corruption with a certain level can represent one distribution.
CIFAR-$10$-C and Tiny ImageNet-C are widely used as benchmarks for test-time adaptation. 
 
To conduct our meta-learning approach, we require a multi-domain training set. 
Thus, we follow the setting in ARM~\cite{zhang2021adaptive} to modify the protocol from \cite{hendrycks2019benchmarking} to construct the multi-domain training set.
Specifically, we apply $56$ distributions on the training sets of CIFAR-$10$ and Tiny ImageNet to form training sets. 
For testing, we apply five corruptions in level $5$ severity on the CIFAR-$10$ and Tiny ImageNet’s test sets to simulate the distribution-shifting streams.
The five distributions in testing have no overlap with the $56$ distributions in training.
We introduce the detailed distribution-shifting setting in the corresponding subsections.

\subsection{Baselines}\label{subsection:baseline}
We compare our method to the following baseline methods. 
\begin{itemize}
  \item {Vanilla}: We train the model by mixing and shuffling all the training data and testing it on the distribution-shifting online streams without adaptation.
  \item Test-time training ({TTT})~\cite{sun2020test} jointly trains the main task and a self-supervised task (such as predicting rotation). In testing, it trains the self-supervised task on the test data to perform the test-time adaptation.
  \item Test-time entropy minimization ({TENT})~\cite{wang2020tent} optimizes the batch normalization layers of the network using the entropy of the network prediction. 
  In test-time, we directly apply the trained {Vanilla} as the initial network.
  \item {ARM}~\cite{zhang2021adaptive} adopts meta-learning to perform the adaptation. In the inner loop updates, it uses three optimization strategies (including CML, LL, and BN). We only compare our methods to {ARM-BN} since the performance of the three strategies is similar.
  \item {NOTE}~\cite{gongnote} replaces the Batch Normalization layers with the proposed Instance-Aware Batch Normalization (IABN). It also uses a reservoir sampling strategy to create a buffer to replay previous data.
  \item {CoTTA}~\cite{wang2022continual} leverages weight-averaged predictions and stochastically restored neurons to adapt to multiple domains continually. In test-time, we directly apply the trained {Vanilla} as the initial network.
\end{itemize}
It is worth noting that {TTT} can adapt to test samples one by one (sample-based), whereas other methods all need to optimize on the entire batch of data (batch-based). Therefore, during the online adaptation, batch-based methods need to make a prediction on each sample first and only update the network after collecting a batch of data.
\begin{figure}
\centering
\includegraphics[trim=0 0 0 0, width=0.48\textwidth]{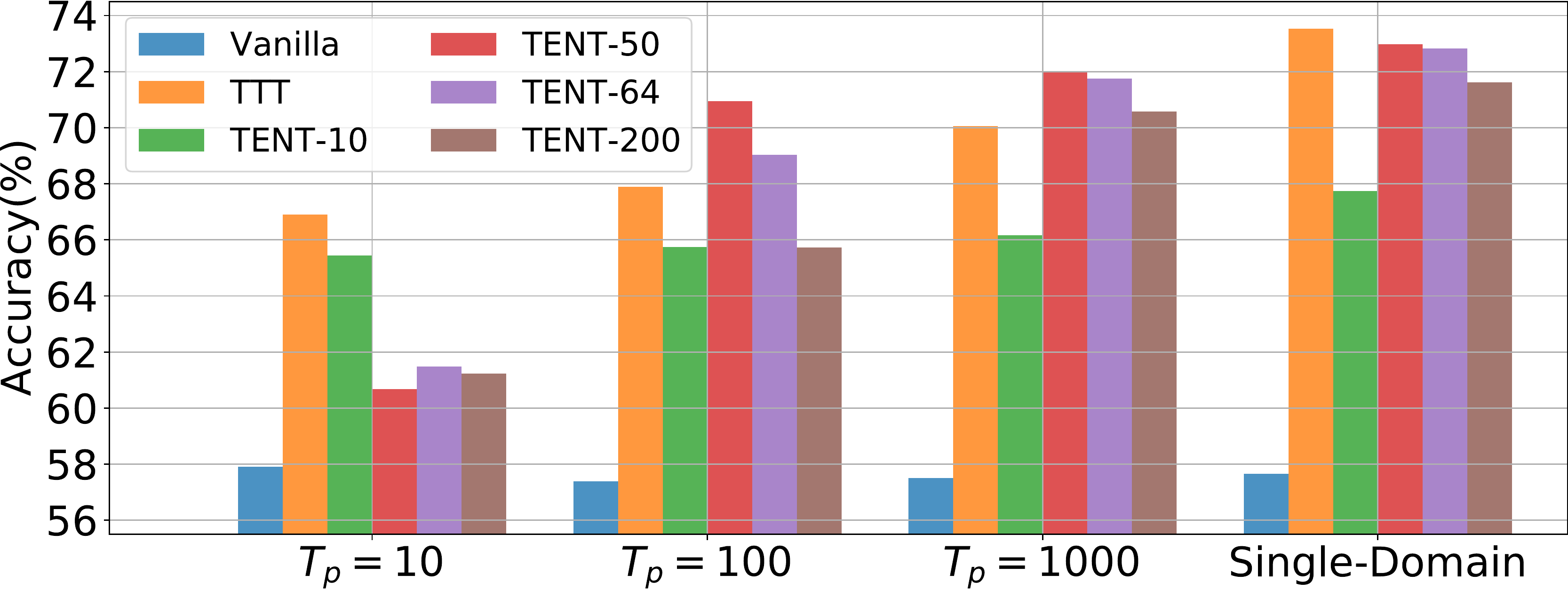}
\caption{Accuracy (\%) of existing methods for distribution-shifting online streams on CIFAR-$10$. }\label{fig:analysis}
\end{figure}

\begin{table*}
  \centering
  \setlength{\tabcolsep}{10pt}
  \definecolor{light-gray}{gray}{0.6}
  \caption{Accuracy ($\%$) for online streams with periodic shifts on CIFAR-$10$-C. Optim. indicates the method optimizes on Batch or Sample. The number after $\pm$ is the $95\%$ confidence interval across five runs with different seeds, followed by the tables shown below.}\label{tab:CIFAR-10}
  \begin{tabular}{ccccccccc} 
    \hline
    \multirow{2}{*}{$T_p$} & \multirow{2}{*}{Method} & \multirow{2}{*}{Optim.} & & \multicolumn{3}{c}{Distribution-shifting domains} & & \multirow{2}{*}{Mean} \\
    \cline{4-8}
    &   &   & Motion  & Impulse    & Spatter  & Jpeg & Elastic  & \\
    \hline
    \multirow{9}{*}{$10$} & Vanilla & - & 67.34\scalebox{0.8}{$\pm$ 0.65} & 60.69\scalebox{0.8}{$\pm$      1.27} & 69.12\scalebox{0.8}{$\pm$      0.53} & 72.19\scalebox{0.8}{$\pm$      0.45} & 69.03\scalebox{0.8}{$\pm$      0.68} & 67.67\scalebox{0.8}{$\pm$      0.37} \\
    & TTT   &    Sample     & 66.59\scalebox{0.8}{$\pm$      0.98} & 63.28\scalebox{0.8}{$\pm$      1.37} & 68.00\scalebox{0.8}{$\pm$      0.49} & {73.44}\scalebox{0.8}{$\pm$      0.53} & 68.78\scalebox{0.8}{$\pm$      0.42} & 68.02\scalebox{0.8}{$\pm$      0.40} \\
    & TENT  &    Batch     & 69.00\scalebox{0.8}{$\pm$      1.41} & 62.52\scalebox{0.8}{$\pm$      1.62} & 67.91\scalebox{0.8}{$\pm$      1.29} & 72.44\scalebox{0.8}{$\pm$      0.68} & 69.54\scalebox{0.8}{$\pm$      0.97} & 68.28\scalebox{0.8}{$\pm$      0.24} \\
    & ARM-BN &   Batch    & 61.87\scalebox{0.8}{$\pm$      0.32} & 54.08\scalebox{0.8}{$\pm$      0.72} & 67.51\scalebox{0.8}{$\pm$      0.48} & 72.03\scalebox{0.8}{$\pm$      0.98} & 67.76\scalebox{0.8}{$\pm$      1.30} & 64.65\scalebox{0.8}{$\pm$      0.28} \\
    & NOTE  &    Batch & 69.27\scalebox{0.8}{$\pm$      0.82} & 64.68\scalebox{0.8}{$\pm$      0.47} & 68.14\scalebox{0.8}{$\pm$      0.95} & 72.67\scalebox{0.8}{$\pm$      0.73} & 69.49\scalebox{0.8}{$\pm$      1.05} & 68.85\scalebox{0.8}{$\pm$      0.42}  \\
    & CoTTA & Batch & 67.92\scalebox{0.8}{$\pm$      1.59} & 61.93\scalebox{0.8}{$\pm$      0.72} & 66.71\scalebox{0.8}{$\pm$      0.74} & 73.23\scalebox{0.8}{$\pm$      0.39} & 69.79\scalebox{0.8}{$\pm$      0.85} & 67.92\scalebox{0.8}{$\pm$      0.38} \\
    \arrayrulecolor{light-gray}\cline{2-9}\arrayrulecolor{black}
    & Ours w/o Adapt. & - & 66.21\scalebox{0.8}{$\pm$      1.03} & 67.46\scalebox{0.8}{$\pm$      1.30} & 70.77\scalebox{0.8}{$\pm$      1.17} & \textbf{73.47}\scalebox{0.8}{$\pm$      1.01} & 71.52\scalebox{0.8}{$\pm$      0.98} & 69.89\scalebox{0.8}{$\pm$      0.36} \\
    & Ours w/o Seq.   & Sample & 63.91\scalebox{0.8}{$\pm$      1.73} & 63.09\scalebox{0.8}{$\pm$      1.98} & 70.29\scalebox{0.8}{$\pm$      0.76} & 72.32\scalebox{0.8}{$\pm$      0.75} & 69.27\scalebox{0.8}{$\pm$      0.53} & 67.78\scalebox{0.8}{$\pm$      0.32}  \\
    & Ours     &  Sample    & \textbf{71.13}\scalebox{0.8}{$\pm$      0.43} & \textbf{68.65}\scalebox{0.8}{$\pm$      1.77} & \textbf{72.07}\scalebox{0.8}{$\pm$      0.92} &73.34\scalebox{0.8}{$\pm$      0.53} & \textbf{71.95}\scalebox{0.8}{$\pm$      1.07} & \textbf{71.43}\scalebox{0.8}{$\pm$      0.49} \\
    \hline
    \multirow{9}{*}{$100$} & Vanilla& -  & 66.20\scalebox{0.8}{$\pm$      0.42} & 60.93\scalebox{0.8}{$\pm$      0.92} & 69.17\scalebox{0.8}{$\pm$      0.66} & 71.61\scalebox{0.8}{$\pm$      0.55} & 70.93\scalebox{0.8}{$\pm$      0.64} & 67.71\scalebox{0.8}{$\pm$      0.17}\\
    & TTT     &    Sample   & 66.59\scalebox{0.8}{$\pm$      0.90} & 64.25\scalebox{0.8}{$\pm$      1.42} & 67.74\scalebox{0.8}{$\pm$      0.79} & 72.75\scalebox{0.8}{$\pm$      0.56} & 69.04\scalebox{0.8}{$\pm$      0.90} & 68.07\scalebox{0.8}{$\pm$      0.33}\\
    & TENT     &   Batch    & 69.62\scalebox{0.8}{$\pm$      0.88} & 63.46\scalebox{0.8}{$\pm$      1.30} & 69.05\scalebox{0.8}{$\pm$      0.86} & 73.38\scalebox{0.8}{$\pm$      1.02} & 70.52\scalebox{0.8}{$\pm$      0.92} & 69.21\scalebox{0.8}{$\pm$      0.53}\\
    & ARM-BN  &    Batch      & 68.73\scalebox{0.8}{$\pm$      0.72} & 62.15\scalebox{0.8}{$\pm$      1.01} & 69.01\scalebox{0.8}{$\pm$      1.04} & 72.80\scalebox{0.8}{$\pm$      1.26} & 69.69\scalebox{0.8}{$\pm$      1.23} & 68.48\scalebox{0.8}{$\pm$      0.31} \\
    & NOTE &    Batch & 69.86\scalebox{0.8}{$\pm$      0.85} & 64.32\scalebox{0.8}{$\pm$      0.51} & 68.68\scalebox{0.8}{$\pm$      1.54} & 72.47\scalebox{0.8}{$\pm$      1.06} & 69.22\scalebox{0.8}{$\pm$      1.28} & 68.91\scalebox{0.8}{$\pm$      0.45} \\
    & CoTTA & Batch & \textbf{70.80}\scalebox{0.8}{$\pm$      0.27} & 63.06\scalebox{0.8}{$\pm$      0.71} & 69.12\scalebox{0.8}{$\pm$      1.29} & {73.73}\scalebox{0.8}{$\pm$      0.75} & 69.63\scalebox{0.8}{$\pm$      0.75} & 69.27\scalebox{0.8}{$\pm$      0.29}\\
    \arrayrulecolor{light-gray}\cline{2-9}\arrayrulecolor{black}
    & Ours w/o Adapt. & - & 65.80\scalebox{0.8}{$\pm$      1.24} & 66.38\scalebox{0.8}{$\pm$      1.57} & 70.59\scalebox{0.8}{$\pm$      1.62} & \textbf{74.91}\scalebox{0.8}{$\pm$      1.75} & 71.34\scalebox{0.8}{$\pm$      0.96} & 69.80\scalebox{0.8}{$\pm$      0.47} \\
    & Ours w/o Seq.   &  Sample     & 63.99\scalebox{0.8}{$\pm$      1.28} & 63.15\scalebox{0.8}{$\pm$      1.27} & 71.06\scalebox{0.8}{$\pm$      0.58} & 73.49\scalebox{0.8}{$\pm$      0.92} & 69.71\scalebox{0.8}{$\pm$      1.65} & 68.28\scalebox{0.8}{$\pm$      0.44} \\
    & Ours      &  Sample   & 70.73\scalebox{0.8}{$\pm$      0.88} & \textbf{69.33}\scalebox{0.8}{$\pm$      1.28} & \textbf{71.68}\scalebox{0.8}{$\pm$      0.66} & 73.18\scalebox{0.8}{$\pm$      1.08} & \textbf{72.12}\scalebox{0.8}{$\pm$      1.30} & \textbf{71.41}\scalebox{0.8}{$\pm$      0.23}\\
    \hline
    \multirow{9}{*}{$1000$} & Vanilla & -  & 67.01\scalebox{0.8}{$\pm$      0.82} & 61.19\scalebox{0.8}{$\pm$      1.20} & 68.35\scalebox{0.8}{$\pm$      1.31} & 71.95\scalebox{0.8}{$\pm$      0.86} & 69.46\scalebox{0.8}{$\pm$      0.87} & 67.59\scalebox{0.8}{$\pm$      0.27} \\
    & TTT   &    Sample   & 67.27\scalebox{0.8}{$\pm$      1.40} & 64.60\scalebox{0.8}{$\pm$      1.17} & 67.00\scalebox{0.8}{$\pm$      0.46} & {73.21}\scalebox{0.8}{$\pm$      0.53} & 68.78\scalebox{0.8}{$\pm$      0.63} & 68.17\scalebox{0.8}{$\pm$      0.42} \\
    & TENT    &   Batch    & 71.33\scalebox{0.8}{$\pm$      0.58} & 63.12\scalebox{0.8}{$\pm$      1.16} & 69.46\scalebox{0.8}{$\pm$      1.01} & 72.55\scalebox{0.8}{$\pm$      2.03} & 69.19\scalebox{0.8}{$\pm$      0.84} & 69.13\scalebox{0.8}{$\pm$      0.73}\\
    & ARM-BN    &   Batch    & 70.55\scalebox{0.8}{$\pm$      0.65} & 64.24\scalebox{0.8}{$\pm$      0.36} & 69.56\scalebox{0.8}{$\pm$      0.86} & 72.55\scalebox{0.8}{$\pm$      1.21} & 70.89\scalebox{0.8}{$\pm$      0.78} & 69.56\scalebox{0.8}{$\pm$      0.14} \\
    & NOTE &    Batch & 70.48\scalebox{0.8}{$\pm$      0.67} & 61.85\scalebox{0.8}{$\pm$      0.78} & 68.85\scalebox{0.8}{$\pm$      0.67} & 72.88\scalebox{0.8}{$\pm$      1.54} & 69.42\scalebox{0.8}{$\pm$      0.73} & 68.70\scalebox{0.8}{$\pm$      0.24}\\
    & CoTTA & Batch & \textbf{71.72}\scalebox{0.8}{$\pm$      0.84} & 64.00\scalebox{0.8}{$\pm$      0.63} & 69.17\scalebox{0.8}{$\pm$      0.92} & 73.14\scalebox{0.8}{$\pm$      0.91} & 71.30\scalebox{0.8}{$\pm$      1.59} & 69.87\scalebox{0.8}{$\pm$      0.17}\\
    \arrayrulecolor{light-gray}\cline{2-9}\arrayrulecolor{black}
    & Ours w/o Adapt. & - & 66.28\scalebox{0.8}{$\pm$      0.84} & 66.61\scalebox{0.8}{$\pm$      0.76} & 70.35\scalebox{0.8}{$\pm$      0.67} & \textbf{74.30}\scalebox{0.8}{$\pm$      0.77} & 72.04\scalebox{0.8}{$\pm$      0.54} & 69.92\scalebox{0.8}{$\pm$      0.23}\\
    & Ours w/o Seq.     &  Sample   & 64.27\scalebox{0.8}{$\pm$      0.88} & 63.85\scalebox{0.8}{$\pm$      1.14} & 70.87\scalebox{0.8}{$\pm$      0.45} & 73.09\scalebox{0.8}{$\pm$      0.75} & 68.52\scalebox{0.8}{$\pm$      0.63} & 68.12\scalebox{0.8}{$\pm$      0.27}\\
    & Ours      &   Sample   & 71.40\scalebox{0.8}{$\pm$      1.22} & \textbf{68.28}\scalebox{0.8}{$\pm$      0.99} & \textbf{71.44}\scalebox{0.8}{$\pm$      0.86} & 72.51\scalebox{0.8}{$\pm$      1.09} & \textbf{72.52}\scalebox{0.8}{$\pm$      1.04} & \textbf{71.23}\scalebox{0.8}{$\pm$      0.35} \\
    \hline
  \end{tabular}
\end{table*}

\subsection{Limitations of Existing Methods}\label{subsection:limit}
We conduct experiments on two typical TTA methods - TTT and TENT, to demonstrate their limitations in the distribution-shifting setting.
To ensure fairness, we adhere to the original implementations of TTT and TENT except for constructing distribution-shifting test streams.

\textit{Distribution-shifting online streams in testing.} 
We randomly arrange the entire CIFAR-$10$ test set (a total of $1,000$ images) into one stream as the test stream.
To simulate the distribution-shifting process in the stream, we adopt a periodic shift: applying five corruptions periodically on the stream.
In chronological order, the following five corruptions are used: Motion Blur, Impulse Noise, Spatter, Jpeg Compression, and Elastic Transform.
One corruption (i.e., domain) lasts for a period of $T_p$ samples before the subsequent corruption takes over.
We set $T_p=10,100,1000$ as three different experiment settings.
$T_p=10$ represents that the distributions shift rapidly, whereas $T_p=1000$ represents that the shifting is relatively slow.
For comparison, we also evaluate the single-domain stream setting, where the models are evaluated on five CIFAR-$10$ online streams separately under the above five corruptions. We obtain the average accuracy over the five streams.

\textit{Implementation.} 
Both TTT and TENT adopt ResNet-$26$~\cite{he2016deep} as the network.
We use open-source code from TTT~\cite{tttcode} to perform the training on the original CIFAR-$10$ training dataset. 
The online test-time adaptation procedures follow the original papers.
As aforementioned, TENT is a typical batch-based method.
To demonstrate the effect of the batch size in this distribution-shifting setting, we set the batch size as $10$, $50$, $64$, and $200$ in the testing.
We denote them as TENT-$10$, $50$, $64$, and $200$.
We also evaluate Vanilla for comparison.

\textit{Result and analysis.} Fig.~\ref{fig:analysis} illustrates the result. We summarize the observations as follows.
\begin{itemize}
  \item TTT and TENT both suffer from the distribution-shifting setting. Their performance on the distribution-shifting streams is significantly worse than the single-domain setting, especially with $T_p=10,100$.
  \item TENT relies on a large batch size to make the adaptation. 
  Fig.~\ref{fig:analysis} shows TENT-10 performs worse than other batch size settings on most streams.
  \item TTT is more robust than TENT in the fast distribution-shifting setting. 
  With $T_p=10$, TTT performs much better than TENT-$50$, $64$, $200$.
  In addition, the performance drop of TTT from $T_p = 1000$ to $T_p = 10$ is not as drastic as that of TENT.
  We hypothesize that TENT requires $T_p$ to be significantly larger than the batch size, as batch-based methods require data in a batch that comes from the same domain.
  The hypothesis is also supported by the fact that TENT-$200$ performs much worse than TENT-$50$ with $T_p = 100$.
\end{itemize}

The above observations inspire our method development. To ensure robustness to fast distribution shifts, we adopt per-sample optimization like TTT. To continually  adapt multiple distributions in testing, we directly learn the distribution-shifting online streams in the inner loop of the meta-learning framework.

\subsection{Periodic Distribution Shifts}\label{subsection:periodic}
This subsection compares our method to state-of-the-art methods on the CIFAR-$10$-C and Tiny ImageNet-C datasets under periodic distribution shifts.
We use the same periodic shifting strategy to construct online streams in testing as in Section~\ref{subsection:limit}, where we set $T_p$ as $10,100,1000$.
We follow ARM to construct the multi-domain training set as aforementioned in Section~\ref{subsection:dataset}.

\textit{Implementation.}
Our architecture and hyper-parameters are consistent across all experiments in this subsection.
We adopt a plain network that has three convolutional layers with $128$ $5\times5$ filters and two fully connected layers. 
The network is the same as the one used in ARM, which is referred to as ConvNet.
The only difference is the normalization layer; we adopt Group Normalization (GN), whereas ARM uses Batch Normalization (BN).
The first two convolutional layers of ConvNet build the feature extractor, and the remaining is the supervised branch.
The self-supervised branch consists of two fully connected layers and a replica of the third convolutional layer of ConvNet.
In the inner loop, we sample $D=3$ domains from $\mathcal{P}^S$ and randomly select $K=5$ samples from each domain.
The initial inner loop learning rate $\alpha$ is set as $0.003$. 
The support set collects additional $D^\prime=20$ domains, resulting in a total of $N=35$ samples.
We use SGD with $\gamma=0.01$ as the optimizer in the outer loop.
The training runs for 100 epochs $\alpha$ and $\gamma$ drop to $0.0003$ and $0.001$ after the $80^\text{th}$ epoch.
We set the test-time learning rate $\beta$ as $0.0003$ to align with the final $\alpha$.

To ensure fairness, we train the ConvNet model from scratch for $100$ epochs for all the baselines.  The only difference is that all the batch-based methods use BN as the normalization layer. During online testing, we set the batch size as 64 for all the batch-based methods. 
It is worth noting that our method shares the same network structure and training data as ARM in the CIFAR-$10$-C experiment.
Thus, we directly use the model trained on ARM's released code~\cite{armcode} to evaluate ARM-BN on our test data.

\textit{Result on CIFAR-10-C.} 
Table~\ref{tab:CIFAR-10} reports the accuracy for all the testing streams under the distribution-shifting setting. 
We also provide the accuracy of each domain for breakdown comprehension. 
Our method achieves the best performance on all the streams and most domains.
The setting of $T_p=10$ produces the most significant gain, where our method outperforms all the competing methods by over $2.5$ points.
Under this setting, all the batch-based methods suffer from fast distribution shifts.
Particularly, ARM-BN even performs worse than Vanilla.
As a sample-based method, TTT only slightly outperforms Vanilla across different settings.
Compared to TTT, ours can quickly adapt to different domains thanks to the meta-learning process.
One interesting thing is other methods commonly perform better as $T_p$ increases, whereas our performance drops slightly. 
This could be because the setting used when constructing inner loop streams (i.e., $k=5$) is closer to $T_p=10$ than $T_p=100, 1000$.
Nevertheless, with $T_p=100, 1000$, ours still outperformed the recent advanced work CoTTA by $2.1$ and $1.4$ points, respectively.

\textit{Ablation on the adaptation in testing.}
We evaluate the performance of our proposed meta-training procedure without incorporating the test-time updates, referred to as ``Ours w/o Adapt." in Table~\ref{tab:CIFAR-10}. The mean accuracy obtained through solo meta-training is $69.89$, $69.80$ and $69.92$ for $T_p = 10, 100,$ and $1000$, respectively. Although the performance is lower than the final method by $1.54$, $1.61$, and $1.31$ points, it is still significantly better than Vanilla. This highlights the importance of both the meta-training procedure and the test-time updates in achieving optimal performance.

\begin{figure}
  \centering
  \includegraphics[trim=0 0 0 0, width=0.48\textwidth]{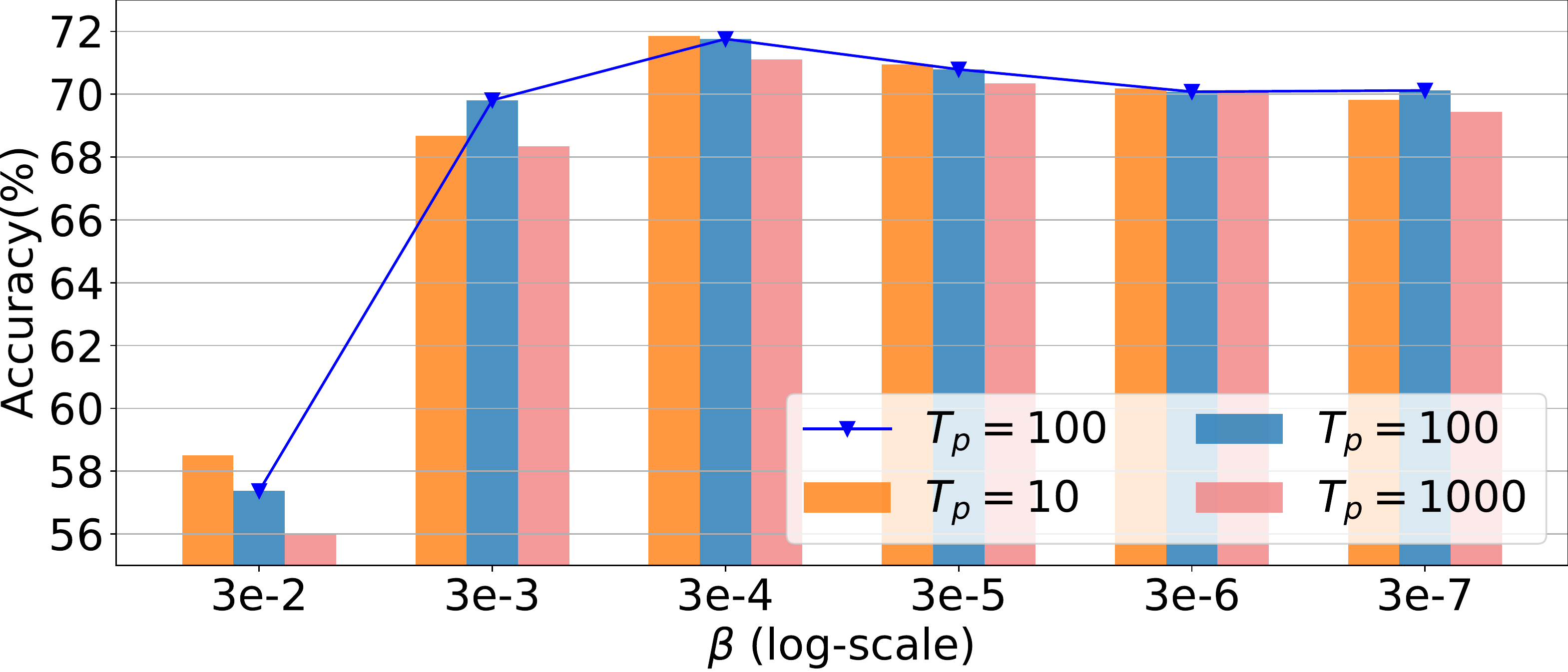}
  \caption{The effect of the test-time learning rate $\beta$. For better visualization, the line chart only operates with $T_p=100$.}\label{fig:abla_lr}
  \end{figure}

We further explore the effect of the test-time learning rate $\beta$ in test-time adaptation.
As aforementioned, we set $\beta$ to be consistent with the final inner loop learning rate $\alpha$. However, it is important to note that different $\beta$ can lead to varying performance outcomes. Fig.~\ref{fig:abla_lr} presents the mean accuracy of a meta-trained model under various $\beta$ in testing. The trends of the mean accuracy across different $\beta$ remain consistent across the three settings of $T_p$. 
From Fig.~\ref{fig:abla_lr}, we can see that a high learning rate severely degraded the performance, while the optimal performance was achieved using a learning rate of $0.0003$, which corresponds to the final $\alpha$ in training. As $\beta$ decreases, performance gradually declines, eventually converging to the scenario with $\beta=0$, which is equivalent to ``Ours w/o Adapt.".

\textit{Ablation on learning from streams.}
Our approach sequentially updates model parameters $\theta_{0}$ in the inner loop.
To investigate how sequential updates can help the performance, we design the following ablation experiment: Given a stream $\mathcal{T}$ with length $L$ in the inner loop, we treat $\mathcal{T}$ as a batch of data. 
Specifically, we compute $\sum_{t=1}^{L}l_{s}(f(x_t;\theta_{0}))$ to replace lines 4-7 of \Cref{algorithm:framework}.
This method is referred to as ``Ours w/o Seq." in Table~\ref{tab:CIFAR-10}.
According to Table~\ref{tab:CIFAR-10}, ``Ours w/o Seq." performs similarly to TTT, much worse than ours.
The result demonstrates the effectiveness of sequential updates in the inner loop.

\textit{Ablation on the support set construction.}
We then analyze the construction of the support set, as shown in Table~\ref{tab:abl_support}. 
Our approach employs self-supervised learning to train samples from $D$ domains in the inner loop, and aims to leverage labeled data to supervise the learning of these domains in the outer loop.
The most straightforward approach to achieving this is by reusing the data from the inner loop as the support set.
However, as shown in the first entry of Table~\ref{tab:abl_support}, this yields a collapse of the model. 
Therefore, we construct the support set by resampling the data points from the $D$ domains, as demonstrated in the second entry, which improves the model performance significantly.
The result aligns with the original MAML algorithm, where the inner loop and support set data are kept distinct.
Even though the inner loop of our algorithm relies on self-supervised learning, it is still necessary to prevent the outer loop from reusing the inner loop data.
In order to increase the model generalization ability to additional domains, we further sample $D^\prime=20$ new domains in the support set. From the third entry in Table~\ref{tab:abl_support}, we can see that the additional samples lead to better performance.

    \begin{table}[t!]
      \centering
      \caption{Ablation on the support set construction.}\label{tab:abl_support}
      \setlength{\tabcolsep}{6pt}
      \begin{tabular}{cccccc}
        \hline
        Reuse/Resamp & Add. &  $T_p=10$ & $T_p=100$ & $T_p=1000$ \\
        \hline
        Reuse & - & 50.24\scalebox{0.8}{$\pm$      0.23} & 50.22\scalebox{0.8}{$\pm$      0.30} & 50.24\scalebox{0.8}{$\pm$      0.20}  \\

        Resample & -    & 70.51\scalebox{0.8}{$\pm$      0.16} & 70.39\scalebox{0.8}{$\pm$      0.37} & 70.01\scalebox{0.8}{$\pm$      0.40}   \\
        Resample & \checkmark    &71.43\scalebox{0.8}{$\pm$      0.49} & 71.41\scalebox{0.8}{$\pm$      0.23}& 71.23\scalebox{0.8}{$\pm$      0.35} \\
        \hline
      \end{tabular}
    \end{table}

    \begin{table}[!t]
      \centering
      \caption{Accuracy ($\%$) for distribution-shifting online streams on Tiny ImageNet-C.}\label{tab:imagenet}
      \setlength{\tabcolsep}{12pt}
      \begin{tabular}{cccc}
        \hline
        Method & $T_p=10$ & $T_p=100$ & $T_p=1000$ \\
        \hline
        Vanilla &16.74\scalebox{0.8}{$\pm$      0.39}&16.87\scalebox{0.8}{$\pm$      0.26}&16.78\scalebox{0.8}{$\pm$      0.33}  \\
         TTT    &19.13\scalebox{0.8}{$\pm$      0.18}&18.92\scalebox{0.8}{$\pm$      0.31}&18.96\scalebox{0.8}{$\pm$      0.34}   \\
         TENT   &17.21\scalebox{0.8}{$\pm$      0.22}&18.53\scalebox{0.8}{$\pm$      0.51} &19.00\scalebox{0.8}{$\pm$      0.13}   \\
         ARM-BN   & 12.12\scalebox{0.8}{$\pm$      0.20} & 18.16\scalebox{0.8}{$\pm$      0.25}& 19.92\scalebox{0.8}{$\pm$      0.21}  \\
         NOTE & 15.45\scalebox{0.8}{$\pm$      0.12} & 15.25\scalebox{0.8}{$\pm$      0.10} & 15.25\scalebox{0.8}{$\pm$      0.16}\\
        CoTTA &17.82\scalebox{0.8}{$\pm$      0.34}& 19.45\scalebox{0.8}{$\pm$      0.20}&19.74\scalebox{0.8}{$\pm$      0.16} \\
        Ours     &\textbf{20.34}\scalebox{0.8}{$\pm$      0.17}& \textbf{20.23}\scalebox{0.8}{$\pm$      0.30}& \textbf{20.08}\scalebox{0.8}{$\pm$      0.27}  \\
        \hline
      \end{tabular}
    \end{table}

\textit{Result on Tiny ImageNet-C.}
We also performed the same set of experiments on the Tiny ImageNet-C dataset. 
Table~\ref{tab:imagenet} reports the mean accuracy for each distribution-shifting online stream.
The results are mostly consistent with CIFAR-$10$-C experiments; 
our method outperforms other competing methods under all three settings. 
That suggests that the advantage of our method is generalizable.

\subsection{Randomized Distribution Shifts}\label{subsection:random_domain}
This subsection evaluates our method in an extreme scenario where the subsequent distribution in a stream is randomly sampled from $\mathcal{P}^T$ instead of following a pre-determined order. 
In this setting, \textit{the duration of each distribution is uncertain} because continuously sampling two identical distributions will extend the duration.
The most extreme case involves the online stream distribution randomly shifting with $T_p=1$, meaning that each frame in the stream is sampled from a randomly selected domain, and the duration of each distribution can vary. For example, if the same domain is consecutively sampled $N$ times, the domain duration will be $N$ samples.
All the trained models here are identical to the ones in Section~\ref{subsection:periodic}, and other testing settings remain the same.

Similar to the experiment with periodic distribution shifts, our method performs the best overall on both CIFAR-$10$-C and Tiny ImageNet-C, as can be seen in Table~\ref{tab:random_shift}. The advantage of our method is more apparent when the distributions shift rapidly (i.e., $T_p=1$ and $T_p=10$).
This experiment shows that our method can perform well under any distribution-shift strategy, even when the shift is completely random.
Consistent with our previous experiments, all of the competing methods suffer from the settings of $T_p=1$ and $T_p=10$. In fact, they show minimal improvements over Vanilla or, in some cases, even worse results.

\begin{table}[t!]
  \caption{Experiments on online streams with randomized shifts.}\label{tab:random_shift}
  \setlength{\tabcolsep}{5pt}
    \begin{subtable}[H]{0.474\textwidth}
      \centering
      \caption{CIFAR-$10$-C}
      \begin{tabular}{ccccc}
        \hline
        Method & $T_p=1$ & $T_p=10$ & $T_p=100$ & $T_p=1000$ \\
        \hline
        Vanilla &67.47\scalebox{0.8}{$\pm$      0.61}&  67.88\scalebox{0.8}{$\pm$      0.22} & 67.52\scalebox{0.8}{$\pm$      0.27} & 67.53\scalebox{0.8}{$\pm$      0.97}  \\
         TTT     &67.81\scalebox{0.8}{$\pm$      0.27}& 67.94\scalebox{0.8}{$\pm$      0.19} & 67.46\scalebox{0.8}{$\pm$      0.34} & 68.01\scalebox{0.8}{$\pm$      1.59}   \\
         TENT   &68.02\scalebox{0.8}{$\pm$      0.66} & 68.27\scalebox{0.8}{$\pm$      0.36}&  68.96\scalebox{0.8}{$\pm$      0.65} & 70.43\scalebox{0.8}{$\pm$      1.09}   \\
         ARM-BN   &64.85\scalebox{0.8}{$\pm$      0.33} & 65.28\scalebox{0.8}{$\pm$      0.40} & 68.51\scalebox{0.8}{$\pm$      0.51} & 68.83\scalebox{0.8}{$\pm$      2.17}   \\
          NOTE  & 68.63\scalebox{0.8}{$\pm$      0.24} & 69.02\scalebox{0.8}{$\pm$      0.20} & 68.63\scalebox{0.8}{$\pm$      0.50} & 68.49\scalebox{0.8}{$\pm$      1.56} \\
        CoTTA &68.28\scalebox{0.8}{$\pm$      0.56}& 68.44\scalebox{0.8}{$\pm$      0.16}&69.43\scalebox{0.8}{$\pm$      0.46}&70.50\scalebox{0.8}{$\pm$      1.05} \\
        Ours     &\textbf{71.57}\scalebox{0.8}{$\pm$      0.25} & \textbf{71.62}\scalebox{0.8}{$\pm$      0.23}& \textbf{71.48}\scalebox{0.8}{$\pm$      0.28}& \textbf{70.92}\scalebox{0.8}{$\pm$      0.97}  \\
        \hline
      \end{tabular}
    \end{subtable}
    
    \begin{subtable}[H]{0.474\textwidth}
      \centering
      \vspace{0.2cm}
      \caption{Tiny ImageNet-C}
      \begin{tabular}{ccccc}
        \hline
        Method & $T_p=1$ & $T_p=10$ & $T_p=100$ & $T_p=1000$ \\
        \hline
        Vanilla & 16.90\scalebox{0.8}{$\pm$      0.31} &16.72\scalebox{0.8}{$\pm$      0.33}&16.76\scalebox{0.8}{$\pm$      0.48}&15.55\scalebox{0.8}{$\pm$      1.90}  \\
         TTT    & 19.12\scalebox{0.8}{$\pm$      0.31} &18.86\scalebox{0.8}{$\pm$      0.24}&18.97\scalebox{0.8}{$\pm$      0.35}&18.36\scalebox{0.8}{$\pm$      1.11}   \\
         TENT   & 17.25\scalebox{0.8}{$\pm$      0.24} &17.13\scalebox{0.8}{$\pm$      0.31}&  18.51\scalebox{0.8}{$\pm$      0.39} & 18.01\scalebox{0.8}{$\pm$      1.45}   \\
         ARM-BN   & 12.12\scalebox{0.8}{$\pm$      0.51} & 12.75\scalebox{0.8}{$\pm$      0.30} & 18.21\scalebox{0.8}{$\pm$      0.67} & \textbf{20.17}\scalebox{0.8}{$\pm$      0.66}   \\
         NOTE & 15.42\scalebox{0.8}{$\pm$      0.12} & 15.18\scalebox{0.8}{$\pm$      0.28}  & 15.51\scalebox{0.8}{$\pm$      0.32} & 15.88\scalebox{0.8}{$\pm$      0.97} \\
        CoTTA & 18.17\scalebox{0.8}{$\pm$      0.13} &17.92\scalebox{0.8}{$\pm$      0.17}&19.33\scalebox{0.8}{$\pm$      0.51}&19.00\scalebox{0.8}{$\pm$      1.35}\\
        Ours  & \textbf{20.37}\scalebox{0.8}{$\pm$      0.08}  &\textbf{20.10}\scalebox{0.8}{$\pm$      0.38}& \textbf{20.23}\scalebox{0.8}{$\pm$      0.61}& 19.41\scalebox{0.8}{$\pm$      1.46}  \\
        \hline
      \end{tabular}
    \end{subtable}
\end{table}
\begin{table}[t!]
  \centering
    \caption{Comparison on CamVid.}\label{tab:seg}
    \setlength{\tabcolsep}{7pt}
  \begin{tabular}{ccccccc}
    \hline
       Method & Vanilla & TENT & NOTE & CoTTA & Ours  \\

      \hline
      mIOU ($\%$)  & 63.48  & 62.02 & 63.97 &64.21 & \textbf{66.67}   \\
      \hline
  \end{tabular}
  \end{table}

\subsection{Video Semantic Segmentation}\label{subsection:video_seg}
 Last, we compare ours with Vanilla, TENT, NOTE, and CoTTA on a practical problem - video semantic segmentation. 

 \textit{Dataset.} We conduct experiments on a driving scene understanding dataset CamVid~\cite{brostow2009semantic}, which contains five video sequences. 
We follow the setting in \cite{wang2021temporal} to split the dataset into training and validation sets.
The validation set is one video sequence containing 101 frames.
The environment of the video frames is constantly changing, resulting in the distribution shift over time.
We arrange the validation set as the test stream.
Following previous work in semantic segmentation~\cite{zhao2017pyramid, long2015fully, chen2017rethinking}, we use Mean Intersection over Union (mIOU) as the metric.

\textit{Implementation.} Ours and Vanilla both adopt PSPNet~\cite{zhao2017pyramid} equipped with ResNet-$50$~\cite{he2016deep} as the network.
To simplify the problem and make the comparison fair, neither uses the Aux branch of PSPNet. 
In our approach, the self-supervised branch consists of one fully connected layer and a replica of the fourth ResLayer in ResNet-$50$.
In the inner loop, we sample four successive frames of one video as one stream $\mathcal{T}$. 
The outer loop combines four randomly sampled frames with the four in the inner loop to form the support set. 
We train $200$ epochs with $\alpha=0.001,\gamma=0.01$.
For Vanilla, we use the open-source code from PSPNet~\cite{semseg2019} to train the Vanilla model for $200$ epochs. 
All the competing methods set the batch size as $4$ in the online testing.

\textit{Result.} Table~\ref{tab:seg} reports the result. 
Compared with Vanilla, TENT does not perform well, NOTE and CoTTA only introduce a slight gain.
This could be because the three batch-based methods suffer from the small batch size setting in this video segmentation task.
Our method outperforms Vanilla and CoTTA by $3.19$ and $2.46$ mIOU.
This substantial improvement demonstrates that our method can be applied in real-world scenarios.

\section{Conclusion}
We presented a novel meta-learning approach for test-time adaptation on distribution-shifting online streams. Our approach explicitly learns how to adapt to these streams by training a meta-learner to quickly adapt to new domains encountered during testing. Our extensive experiments show that our approach outperforms existing state-of-the-art methods in multiple distribution-shifting settings. 
In the future, we plan to apply our method to other practical tasks, such as human body orientation estimation~\cite{wu2020mebow}. We hope that our work can provide a foundation for building more robust and adaptable AI systems that can perform well in dynamic and changing environments.

{\it Acknowledgments:} C.W. and J.Z.W. were supported in part by a generous gift from the Amazon Research Awards program. 
They utilized the Extreme Science and Engineering Discovery Environment
and Advanced Cyberinfrastructure Coordination Ecosystem: Services \& Support. supported by NSF.
We acknowledge Xianfeng Tang, Huaxiu Yao, and Jianbo Ye for their helpful discussions.

\bibliography{source}
\bibliographystyle{ieee}

\end{document}